\documentclass{llncs}

\usepackage{graphicx,times,amsmath} 
\usepackage{url}
\usepackage{graphicx}
\usepackage{amsfonts}
\usepackage{wasysym}
\usepackage{latexsym}
\usepackage{color}
\usepackage{multirow}
\usepackage{amssymb}
\usepackage{epstopdf}

\usepackage{array}
\usepackage{multirow} 
\usepackage{verbatim}
\usepackage{amstext}
\usepackage{amsmath}
\usepackage{rotating}
\usepackage{graphicx}
\usepackage{amssymb}
\usepackage{subfigure}
\usepackage[svgnames]{xcolor}
\usepackage{float} 
\usepackage[ruled,vlined]{algorithm2e}
\usepackage{multirow}
\usepackage{url}
\usepackage{colortbl}
\usepackage{caption}

\newcommand{\argmax}[1]{\underset{#1}{\operatorname{arg}\,\operatorname{max}}\;}
\newcommand{\nomAlgo}{NeuralBandit1}
\newcommand{\nomAlgoa}{NeuralBandit2}
\newcommand{\nomAlgob}{NeuralBandit3}

\begin{document}

\title {A Neural Networks Committee for the Contextual Bandit Problem}





\author{Robin Allesiardo\inst{1}$^{,}$\inst{2} \and Rapha\"{e}l F\'{e}raud\inst{2} \and Djallel Bouneffouf\inst{2}}

\authorrunning{Allesiardo et al.} 

\institute{TAO - INRIA, CNRS, University of Paris-Sud, 91405 Orsay (France) 
\and Orange Labs, 2 av. Pierre Marzin, 22300 Lannion (France)}

\maketitle 


\begin{abstract}
This paper presents a new contextual bandit algorithm, NeuralBandit, which does not need hypothesis on stationarity of contexts and rewards.
Several neural networks are trained to modelize the value of rewards knowing the context.
Two variants, based on multi-experts approach, are proposed to choose online the parameters of multi-layer perceptrons.
The proposed algorithms are successfully tested on a large dataset with and without stationarity of rewards.

\end{abstract}


\section{Introduction}
\paragraph{}{

In online decision problems such as online advertising or marketing optimization, a decision algorithm must select amoung several actions. Each of these options is associated with side information (profile or context) and the reward feedback is limited to the chosen option. For example, in online advertising, a visitor queries a web page; a request with the context (web page address, cookies, customer profile, etc.) is send to the server; the server sends an ad which is displayed on the page. If the visitor clicks on the ad the server receives a reward. The server must trade-off between the explorations of new ads and the exploitation of known ads.
Moreover, in an actual applications, both rewards and data distributions can change with time. For instance, the display of a new ad can change the probability of clicks of all ads, the content of a web page can change over time. Robustness to non-stationarity is thus strongly recommended. 
}



\section{Previous work}
\label{sec:travauxPrecedents}
\paragraph{}{


The multi-armed bandit problem is a model of exploration and exploitation where one player gets to pick within a finite set of decisions the one which maximizes the cumulated reward.
This problem has been extensively studied. Optimal solutions have been provided using a stochastic formulation ~\cite {LR85,UCB}, a Bayesian formulation ~\cite {T33,kaufmann}, or using an adversarial formulation ~\cite{AuerC98,AuerCFS02}. 
Variants of the initial problem were introduced due to practical constraints (appearance of a new advertisment after the beginning of learning~\cite{KleinbergNS08,ChakrabartiKRU08}, fixed number of contractual page views ~\cite{FeraudU12,FeraudU13}). However these approaches do not take into account the context while the arm's performance may be correlated therewith.

A naive solution to the contextual bandit problem is to allocate one bandit problem for each context.
A tree structured bandit such as X-armed bandits ~\cite{BubeckMSS08} or a UCT variant ~\cite{KocsisS06} can be used to explore and exploit a tree structure of contextual variables to find the leaves which provide the highest rewards ~\cite{GaudelS10}.
The combinatorial aspect of these approaches limits their use to small context size.
Seldin et al ~\cite{SeldinALSO11} modelize the contexts by state sets, which are associated with bandit problems. 
Without prior knowledge of the contexts, it is necessary to use a state per context, which is equivalent to the naive approach. 
Dud\' ik et al ~\cite{dudik11} propose an algorithm of policies elimination. The performance of this algorithm depends on the presence of a good policy in the set.
The epoch-greedy algorithm ~\cite{LangfordZ07} alternates exploration then exploitation. 
During exploration, the arms to play are randomly drawn to collect an unbiased training set. Then, this set is used to train a classifier which will be used for exploitation during the next cycle.
The nature of the classifier remains to be defined for concrete use. In LINUCB ~\cite{Li2010,ChuLRS11} and in Contextual Thompson Sampling (CTS)~\cite{Agrawal2013}, the authors assume a linear dependency between the expected reward of an action and its context and model the representation space using a set of linear predictors. Banditron ~\cite{Kakade2008} uses a perceptron 
per action to recognize rewarded contexts.
Furthermore, these algorithms assume that the data and the rewards are drawn from stationary distribution, which limits their practical use. The EXP4 algorithm \cite{AuerCFS02} selects the best arm from $N$ experts advices (probability vectors). Exploring the link between the rewards of each arm and the context is delegated to experts. Unlike the previous algorithms, the rewards are assumed to be chosen in advance by an adversary. Thus, this algorithm can be applied to non-stationnary data.

At first we will formalize the contextual bandit problem and will propose a first algorithm: {\nomAlgo}. Inspired by Banditron ~\cite{Kakade2008} it estimates the probabilities of rewards by using neural networks in order to be free of the hypothesis of linear separability of the data. 
Neural networks are universal approximators ~\cite{Hornik89}. They are used in reinforcement learning ~\cite{Tesauro02,Mnih2013} and can estimate accurately the probabilities of rewards within actual and complex problems. In addition, the stochastic gradient achieves good performances in terms of convergence to the point of best generalization ~\cite{bottou2004} and has the advantage of learning online. In seeking to reach a local minimum, the stochastic gradient can deal with non-stationarity. This will result in a change of the cost function landscape over the time. If this landscape changes at a reasonable speed, the algorithm will continue the descent to a new local minimum. The main issue raised by the use of multilayer perceptrons remains the online setting of various parameters such as the number of hidden neurons, the value of the learning step or the initalization of the weight seed. We propose two advanced versions of the algorithm {\nomAlgo} to adjust these settings using adversarial bandits that seek, among several models initialized with different parameters, the best one. We conclude by comparing these different approaches to the state-of-the-art on stationnary and on non-stationnary data.

}

\section{Our algorithm: {\nomAlgo}}
\begin{definition}[Contextual bandit]
{Let $x_t\in X$ be a contextual vector and $(y_{t,1},...,y_{t,K})$ a vector of rewards associated with the arm $k\in [K]=\{ 1,...,K\} $ and $((x_1,\textbf{y}_1),...,(x_T,\textbf{y}_T))$ the sequence of contexts and rewards. The sequence can be drawn from a stochastic process or chosen in advance by an adversary. At each round $t<T$, the context $x_t$ is announced. The player, who aims to maximize his cumulated rewards, chooses an arm $k_t$. The reward $y_{t,k}$ of the played arm, and only this one, is revealed.
}
\end{definition}

\begin{definition}[Cumulated regret]
{
Let $H:X\rightarrow [K]$ be a set of hypothesis , $h_t \in H$ a hypothesis computed by the algorithm $A$ at the round $t$ and $h^{*}_{t}=\underset{h_t \in H}{\operatorname{argmax}}\ y_{t,h_{t}(x_t)}$ the optimal hypothesis at the same round. The cumulated regret is: 
\begin{equation*}
R(A) = \sum ^{T}_{t=1} {y_{t,h^{*}_{t}(x_t)}- y_{t,h_t(x_t)}}
\end{equation*}
The purpose of a contextual bandit algorithm is to minimize the cumulative regret. 
}
\end{definition}

{Each action $k$ is associated with a neural network with one hidden layer which learns the probability of reward for an action knowing its context. We choose this modelization rather than one neural network with as many outputs as actions to be able to add or remove actions easier.

Let $K$ be the number of actions, $C$ the number of neurons of each hidden layers and ${\mathbf{N}}_{t}^{k}:X\rightarrow Y$ the function associating a context $x_t$ to the output of the neural network corresponding to the action $k$ at the round $t$. $N$ denotes the number of connections of each network with $N=dim(X)C+C$. To simplify the notations, we place the set of connections in the matrix $W_t$ of size $K \times N$. Thus, each row of the matrix $W_t$ contains the weight of a network. $\Delta _t$ is the matrix of size $K \times N$ containing the update of each weight between rounds $t$ and $t+1$. The update equation is:

\begin{equation*}
W_{t+1} = W_{t} + \Delta _t
\end{equation*}

The backpropagation algorithm ~\cite{Rumelhart86} allows calculating the gradient of the error for each neuron from the last to the first layer by minimizing a cost function. Here, we use the quadratic error function and a sigmoid activation function.

Let $\lambda$ be the learning step, $\hat{x}_{t}^{n,k}$ the input associated with the connection $n$ in the network $k$, $\delta_{t}^{n,k}$ the gradient of the cost function at round $t$ for the neuron having as input the connection $n$ in the network $k$ and $\Delta _t^{n,k}$ the value corresponding to the index $(n,k)$ of the matrix $\Delta _t$. When the reward of an arm is known, we can compute:

\begin{equation*}
\Delta _t^{n,k} = \lambda \hat{x}_{t}^{n,k} \delta_{t}^{n,k}
\end{equation*}

In the case of partial information, only the reward of the arm $k_t$ is available. To learn the best action to play, an approach consists of a first exploration phase, where each action is played the same number of times in order to train a model, and then an exploitation phase where the obtained model is used. Thus, the estimator would not be biased on the most played action. However, this approach would have abysmal performances in case of non-stationary data. We choose to use an exploration factor $\gamma$, constant over time, allowing continuing the update of the model in the case of non-stationary data. The probability of playing the action $k$ at round $t$ knowing that $\hat{k}_t$ is the arm with the highest reward prediction is:

\begin{equation*}
\mathbf{P}_t(k) = (1-\gamma) \mathbf{1}[k=\hat{k}_t] + \frac{\gamma}{K}
\end{equation*}

We propose a new update rule taking into account the exploration factor:

\begin{equation}
W_{t+1} = W_{t} + \tilde{\Delta}_t \text { , }
\label{updateNN}
\end{equation}

\begin{equation*}
\text {with }\tilde{\Delta} _t^{n,k} = \frac{ \lambda \hat{x}_{t}^{n,k} \delta_{t}^{n,k} \mathbf{1}[\hat{k}_t=k]}{\mathbf{P}_t(k)}
\end{equation*}

\begin{theorem}
The expected value of $\tilde{\Delta} _t^{n,k}$ is $ \Delta _t^{n,k}$.
\end{theorem}

\paragraph{The proof is straightforward:}
{
\begin{equation*}
\begin{array}{rcl}
\mathbf{E}[\tilde{\Delta} _t^{n,k}] &=& \sum_{k=1}^{K} \mathbf{P}_t(k)(\frac{ \lambda \hat{x}_{t}^{n,k} \delta_{t}^{n,k} \mathbf{1}[\hat{k}_t=k]}{\mathbf{P}_t(k)})\\ &=& \lambda \hat{x}_{t}^{n,k} \delta_{t}^{n,k} \\ &=& \Delta _t^{n,k}
\end{array}
\end{equation*}\qed
}

The proposed algorithm, {\nomAlgo}, can adapt to non-stationarity by continuing to learn over time, while achieving the same expected result (in the case of stationary data) as a model trained in a first phase of exploration. 

\begin{algorithm}
\label{NB1}
\DontPrintSemicolon
\KwData{$\gamma \in [0,0.5]$ et $\lambda \in ]0,1]$}
\Begin{
Initialize $W_1 \in ]-0.5,0.5[^{N \times K}$\;
\For{$t = 1,2,...,T$}{
Context $x_t$ is revealed\;
$\hat{k}_t = \argmax{k\in[K]}\mathbf{N}_{t}^{k}(x_t)$\;
$\forall k \in [K] \text{ on a } \mathbf{P}_t(k) = (1-\gamma) \mathbf{1}[k=\hat{k}_t] + \frac{\gamma}{K}$\;
$\tilde{k}_t$ is drawn from $\mathbf{P}_t$\;
$\tilde{k}_t$ is predicted and $y_{t,\tilde{k}_t}$ is revealed\;
Compute $\tilde{\Delta} _t$ such as $\tilde{\Delta} _t^{n,k} = \frac{ \lambda \hat{x}_{t}^{n,k} \delta_{t}^{n,k} \mathbf{1}[k_t=k]}{\mathbf{P}_t(k)} $\;
$W_{t+1} = W_{t} + \tilde{\Delta}_t$\;
}

}

\caption{{\nomAlgo} \label{IR}}
\label{fig:NB1}
\end{algorithm}

\section{Models selection with adversarial bandit}
\paragraph{}
{Performances of neural networks are influenced by several parameters such as the learning step, the number of hidden layers, their size, and the initalization of weights. The multi-layer perceptron corresponding to a set of parameters is called model. Using batch learning, the models selection is done with a validation set. Using online learning, we propose to train the models in parallel and to use the adversarial bandit algorithm EXP3 ~\cite{AuerC98,AuerCFS02} to choose the best model. The choice of an adversarial bandit algorithm is justified by the fact that the performance of each model changes overtime due to the learning itself or due to the non-stationarity of the data.}


\subsubsection*{Exp3}
{ Let $\gamma_\text{model} \in [0,1]$ be an exploration parameter, $w_t=(w^{1}_t,...,w^{M}_t)$ a weight vector, where each of its coordinate is initialized to $1$, and $M$ the number of models. Let $m$ be the model chosen at time $t$, and $y_{t,m}$ be the obtained reward. The probability to choose $m$ at round $t$ is:
}
\begin{equation}
\mathbf{P_\text{model}}_t(m) = (1-\gamma_\text{model}) \frac{w^{m}_t}{\sum_{i=1}^{M} w_{t}^{i}} + \frac{\gamma_\text{model}}{M}
\label{exp3proba}
\end{equation}

\paragraph{}
{
The weight update equation is: 
}
\begin{equation}
w_{t+1}^{i} = w_{t}^{i}\text{exp}\left(\frac{\gamma_\text{model}\mathbf{1}[i=m]y_{t,m}}{\mathbf{P_\text{model}}_t(i)M}\right)
\label{exp3update}
\end{equation}

\subsubsection{{\nomAlgoa} (see Algorithm~\ref{fig:NB2})}
{ If we consider that a model is an arm, a run of this model can be considered as a sequence of rewards. 
The algorithm takes as inputs a list of $M$ models and a model exploration parameter $\gamma _{\text{model}}$. For each element of the list, one {\nomAlgo} instance is initialized.
Each instance corresponds to an arm. {\sc EXP3} algorithm is used to choose the arms over time. 
After receiving a reward, each neural network corresponding to the played arm is updated, and the weights of {\sc EXP3} are updated.
}
\subsubsection{{\nomAlgob} (see Algorithm~\ref{fig:NB3})}
The use of the algorithm {\nomAlgoa} corresponds to the assumption that there is a model {\nomAlgo} which is the best for all actions.
The algorithm {\nomAlgob} lifts this limitation by associating one {\sc Exp3} per action.


\begin{figure}
\begin{algorithm}[H]
\KwData{$\gamma _{\text{model}} \in [0,0.5]$ and a list of $M$ models parameters}
\Begin{
Initialize $M$ {\nomAlgo} \;
Initialize the EXP3 weight vector $w_0$ with $\forall m\in [M] $ $w_{0}^m=1$\;
\For{$t = 1,2,...,T$}{
Context $x_t$ is revealed\;
$m_t$ is drawn from $\mathbf{P_\text{model}}_t$ ~\eqref{exp3proba}\;
The model $m_t$ choose action $\tilde{k}_t$\;
$\tilde{k}_t$ is predicted and $y_{t,\tilde{k}_t}$ is revealed\;
Update of each network corresponding to the action $\tilde{k}_t$ for each model with ~\eqref{updateNN}\;
Update of the EXP3 weight vector $w_t$ with ~\eqref{exp3update}\; 
}
}

\caption{{\nomAlgoa} \label{IR}}
\label{fig:NB2}
\end{algorithm}

\begin{algorithm}[H]
\KwData{$\gamma \in [0,0.5]$, $\gamma _{\text{model}} \in [0,0.5]$and a list of $M$ model parameters}
\Begin{
Initialize $K$ neural networks per model $m$ \;
Initialize $K$ instance of EXP3\;
\For{$t = 1,2,...,T$}{
Context $x_t$ is revealed\;
\For{$k = 1,2,...,K$}{
$m_{t}^k$ is drawn from $\mathbf{P}_{\text{model}^k_t}$ ~\eqref{exp3proba}\;
Action $k$ is scored $s^k _t=\mathbf{N}_{t}^{m_{t}^k,k}(x_t)$\;
}
$\hat{k}_t = \argmax{k\in[K]}s^k _t$\;
$\forall k \in [K] \text{ on a } \mathbf{P}_t(k) = (1-\gamma) \mathbf{1}[k=\hat{k}_t] + \frac{\gamma}{K}$\;
$\tilde{k}_t$ is drawn from $\mathbf{P}_t$\;
 $\tilde{k}_t$ is predicted and $y_{t,\tilde{k}_t}$ is revealed\;
Update of each network corresponding to the action $\tilde{k}_t$ for each model with ~\eqref{updateNN}\;
Update of each EXP3 weight vectors with ~\eqref{exp3update}\; 
}
}
\caption{{\nomAlgob} \label{IR}}
\label{fig:NB3}
\end{algorithm}

\end{figure}
\paragraph{}{

{\nomAlgob} has greater capacity of expression than {\nomAlgoa} as each action can be associated with different models. However if the best model in {\nomAlgob} exists in {\nomAlgoa}, then {\nomAlgoa} should find this model faster than {\nomAlgob} because it has only one instance of EXP3 with less possibility to update.

\section{Experiments}
\paragraph{}
{
The Forest Cover Type dataset from the UCI Machine Learning Repository is used. It contains 581.000 instances and it is shuffled. We have recoded each continuous variable using equal frequencies into five binary variables and we have recoded each categorical variable into binary variables. We have obtained 94 binary variables for the context, and we have used the 7 target classes as the set of actions. In order to simulate a datastream, the dataset is played in loop. At each round, if the algorithm chooses the right class the reward is 1 or else 0. The cumulated regret is computed from the rewards of an offline algorithm fitting the data with 93\% of classification. The plot of the curves (Figure ~\ref{fig:figcov_all}) are produced by averaging 10 runs of each algorithms with $\gamma=0.005$, $\gamma_{model}=0.1$. Each run began at a random position in the dataset.
The parameters of each model ({\nomAlgo}) are the combination of different sizes of hidden layer $(1,5, 25, 50, 100)$ and different values of $\lambda$ $(0.01,0.1,1)$. 
}
\subsubsection{On stationary data (left part of Figure ~\ref{fig:figcov_all})} Banditron achieves a high cumulated regret (57\% of classification computed on the 100.000 last predictions) and is outperformed by all other contextual bandit algorithms on this dataset. The cumulated regret and classification rate of LinUCB (72\%), CTS (73\%) and {\nomAlgob} (73\%) are similar and their curves tend to be the same. {\nomAlgoa} has the fastest convergence rate and achieves a smaller cumulated regret with 76\% of classification.

\subsubsection{On non-stationary data (right part of Figure ~\ref{fig:figcov_all})}
Non-stationarity is simulated by swapping classes with a circular cycle ($1\rightarrow 2$, $2\rightarrow 3$ ,...,$7\rightarrow 1$) every 500.000 iterations. At each concept drift, curves increase then stabilize. On stationary data LinUCB and CTS achieve a lowest regret than Banditron but can't deal with non-stationarity thus are outperformed by it after the first drift. Banditron converges again near instantanly. Models selection algorithms need between 75.000 and 350.000 in the worst case to stabilize at each drift. {\nomAlgoa} and {\nomAlgob} are more complex models than Banditron but achieve better performances on this non-stationary datastream. {\nomAlgob} seem to be more robust to nonstationarity than {\nomAlgoa} on this dataset. This can be explained by the fact that sometime, one neural network can stay on a bad local minimum. If this append in {\nomAlgoa}, the entire model is penalized while in {\nomAlgob} the algorithm can still use all the other networks.

\begin{figure}
 \centering
 \includegraphics[width=1.0\textwidth]{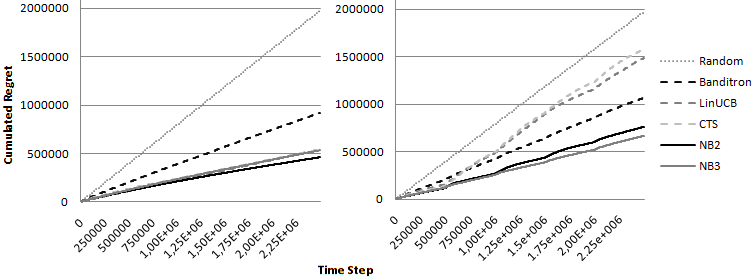}
 \caption{The cumulated regret of different contextual bandit algorithms over time on Forest Cover Type. The left part is on a stationary datastream and the right part on a non-stationary datastream.}
 \label{fig:figcov_all}
\end{figure}
\vspace{-10mm}

\section{Conclusion}
\paragraph{}
{
We introduced a new contextual bandit algorithm {\nomAlgo}. Two variants with models selection {\nomAlgoa} and {\nomAlgob} used an adversarial bandit algorithm to find the best parameters of neural networks. We confronted them to stationary and non-stationary datastream. They achieve a smaller cumulated regret than Banditron, LinUCB and CTS. This approach is successful and has the advantage of being trivially parallelizable. Models differentiation show a significant gain on the Forest Covert Type dataset with nonstationarity. We also showed empirically that our two models selection algorithms are robust to concept drift. These experimental results suggest that neural networks are serious candidates for addressing the issue of contextual bandit. 
However, they are freed from the constraint of linearity at the expense of the bound on the regret, the cost function being not convex. }

\vspace{-2mm}
\bibliographystyle{splncs}
\bibliography{biblio}
\end{document}